# Improving Bangla Linguistics: Advanced LSTM, Bi-LSTM, and Seq2Seq Models for Translating Sylheti to Modern Bangla

Sourav Kumar Das
*Dept. of CSE*
*Daffodil International University*
Dhaka, Bangladesh
sourav15-4588@diu.edu.bd

Md. Julkar Naeen
*Dept. of CSE*
*Daffodil International University*
Dhaka, Bangladesh
naeen15-4578@diu.edu.bd

MD. Jahidul Islam
*Dept. of CSE*
*Daffodil International University*
Dhaka, Bangladesh
jahidul3154@diu.edu.bd

Md. Anisul Haque Sajeeb
*Dept. of CSE*
*Daffodil International University*
Dhaka, Bangladesh
haque15-7533@diu.edu.bd

Narayan Ranjan Chakraborty
*Dept. of CSE*
*Daffodil International University*
Dhaka, Bangladesh
narayan@daffodilvarsity.edu.bd

Mayen Uddin Mojumdar
*Dept. of CSE*
*Daffodil International University*
Dhaka, Bangladesh
mayen.cse@diu.edu.bd

*Abstract*-Bangla or Bengali is the national language of Bangladesh, people from different regions don't talk in proper Bangla. Every division of Bangladesh has its own local language like Sylheti, Chittagong etc. In recent years some papers were published on Bangla language like sentiment analysis, fake news detection and classifications, but a few of them were on Bangla languages. This research is for the local language and this particular paper is on Sylheti language. It presented a comprehensive system using Natural Language Processing or NLP techniques for translating Pure or Modern Bangla to locally spoken Sylheti Bangla language. Total 1200 data used for training 3 models LSTM, Bi-LSTM and Seq2Seq and LSTM scored the best in performance with 89.3% accuracy. The findings of this research may contribute to the growth of Bangla NLP researchers for future more advanced innovations.

## I. Introduction

Bangla is the fifth largest language in terms of native speakers and the sixth largest in terms of the total number of speakers. Bangla is the state language of Bangladesh and the official language of West Bengal, Tripura and Assam, also it's the main language in Andaman Island. Some other states also use this language. More than 28.5 million people around the world use Bangla as their way of communication. In Bangladesh, there are a lot of areas that use Bangla and the local language is different from the main Bangla language. It's quite hard to understand the local language of some particular area. Finally, the research may give us the opportunity to understand the local language of some particular area. The objective of the research is to make sure everyone can understand the local language by using some tools or applications. There mustn't remain any communication gap between those who use Bangla as their primary language. This work can be used in research for other research purposes and even in general translation from local Bangla to Modern Bangla.

Natural language processing (NLP) plays a critical role in this research. Researching this type of resource requires a lot of data to start with to get a good result. Also English to Bangla translation has been done by the researchers using neural networks [1]. A group of researchers had developed resources for Indonesian local language using Indo-BERT [2]. The researchers are focusing on this topic nowadays because it's getting popular among the researchers, also the study is interesting.

Unfortunately, there isn't enough research in the Bangla Local language translation from Local Bangla to Modern Bangla. This study aims to translate Local Bangla language to Modern Bangla for better communication. By doing the research, an app could be developed to do the translation work for better user experience. This study will help the normal people to understand any local language that is available in Bangladesh. Also, this study will help a lot of researchers and the researcher will be inspired to do more research about this topic in the future. Moreover Bangla, mainly local Bangla will be more understandable. The simple goal was to translate Sylheti Bangla to Modern Bangla. For this research, the first priority was to collect the dataset and it was collected from people of Sylhet who can tell their language better than anyone. After that some deep learning models were applied to do the research. Before that for better understanding, some preprocessed techniques were used, so that models can learn better. And it will give a good accuracy.

## II. Literature Review

Compared to the other languages, Bangla has been the least touched language in research specifically in machine learning or artificial intelligence. But recently Bangladeshi researchers are working with Bangla language. But most of them are sentiment analysis, fake news detection. A very little amount, of papers are about the locally spoken Bangla language. Some recent papers relevant to our work are reviewed. Ankon Sarkar et al..[3] provided Bengali texts dataset for sentiment analysis to extract emotions like (Happiness, Sadness, Fear, Anger, Disgust Surprise etc.) from the given dataset using deep learning techniques like LSTM, TF-IDF, BERT. The BERT model outperformed all models and the accuracy was 69.2%. The data is collected from web platforms like social media. The quantity of data is 14,999 in total. There might be some inconsistency with the annotations and for this reason the model might disagree in some cases.

Md. Tofael Ahmed et al. [4] provided a dataset of Bengali and Romanized Bengali for cyberbullying detection using the SVM and Naive Bayes models. Naive Bayes outperformed over all other models and the best accuracy was 84%. The data was divided into 3 parts, 5000 Bangla comments, 7000 Bangla Romanized Bangla, 12000 texts. They collected the data from YouTube.

Md Gulzar Hussain et al. [5] proposed a Bangla fake news detection system from social media using MNB and SVM classifiers. The models worked quite well, SVM with the best accuracy of 96.64%. Count-vectorizer and term frequency-Inverse document frequency vectorizer used for feature extraction and the dataset was collected from different newspapers in Bangla language. 2500 articles were collected for this topic. The success rate can be upgraded by utilizing hybrid-classifiers and deep learning models.

Zobaer et al. [6] presented an automated system that detects Bangla fake news with NLP techniques. Total 50K of data used for this system. SVM, Linear Regression and Random Forest total three models were used. SVM provided the best score in all the traditional linguistic features. Some terms like Precision, Recall and F1-Score of fake news vary in experiments, and the reason was to set 37.47 times less the numbers of fake news than the authentic news.

Shafayet Bin Shabbir Mugdha et al. [7] presented a system that can detect news which are fake using full content or accuracy based on the news headline, performing models like Gaussian Naive Bayes, Extra tree classifier. GNB gave a decent accuracy of 87%. The dataset used in the topic was a novel dataset of Bengali language. This topic could work more effectively by using more meaningful data, so it can generate more attributes.

Kamrus Salehin et al. [8] had studied different text classification for Bangla news classification. By applying five different ML and two neural networks LSTM outperformed every model, its accuracy was 87%. Total 75951 data were used for this topic. Data was collected from Bangla online news portals. Authors didn't use other feature selection and there was no Bangla stemming, lemmatization that had good qualities.

Ovishake Sen et al..[9] recommend a process to recognize Bangla Speech. Total 400 audios had been taped to create this set of data. In the dataset they applied MFCCs, CNNs. The suggested technique is recognized with an accuracy of 97.1%. If linguistic data from all ages and genders were gathered, this would create a better model.

Jinal H. Tailor et al. [10] suggested an approach for spoken digit identification. By applying CNN and MFCC models to analyze the audio clips. By applying a process 98.7% accuracy was gained. Dataset was generated by a total of 8 persons with age between 20-40 and a total 2400 audio clips were recorded by both of these genders. By expanding the dataset size, the authors could improve the accuracy as well as the overall model.

Khaled M. Fouad et al..[11] offered a model for detecting news which appeared to be fake in Arabic language using deep learning. In the work ML and DL model both were applied. SGD Classifier gave a good accuracy of 86.1% among all three datasets in both ML and deep learning. CNN, LSTM(Bi-LSTM), CNN+LSTM were used depending on the dataset. Three datasets were used and they were collected from Arabic news articles.

Nitish Ranjan Bhowmik et al. [12] presented sentiment analysis on Bangla text using supervised ML with the help of a lexicon dictionary. Models like LR, KNN, RF, SVM were applied and the unigram + Proposed model gave the highest accuracy of 85%. There were 2 datasets used in the research and it was founded in GitHub. If more than 50 thousand data were used in the process, accuracy would have been higher.

Rajesh Kumar Das et al. [13] focused on categorizing Bangla sentences into three types of sentences, they are simple, complex and compound sentences. Several models were applied like LSTM, bi-LSTM, Conv1D and hybrid models combining Conv1D and LSTM models. Combined model gave the highest accuracy of 91.17%. Total number of sentences was 3793 and was collected from Facebook pages that are accessible publicly.

Nitish Ranjan Bhowmik et al. [14] proposed a method using deep learning approach with rule, based method (BTSC) for sentiment analysis of Bangla text using LDD which is an extended lexicon data dictionary. Different LSTM models are used in this case and got the best accuracy of 84.18% from BERT-LSTM. This paper also includes fine tuning for presenting detailed analysis of DL models.

Mehedi Hasan et al. [15] presented a topic about multiple Bangla sentence classification using ML and Deep learning algorithms like KNN, RF, DT, SVM, MNB, LSTM, RNN. Among these algorithms RF and DT gave the highest accuracy and it was 89.42% for both. Total 1824 text documents were observed. The dataset was created by hand. The dataset was insufficient and with closed domains.

Abu Sayeed et al. [16] proposed a topic about Bengali Handwritten characters recognition using neural networks. There was a total of 8 dataset and there was more than one lakh data in total. Some of the data were imbalanced too. Authors applied some techniques in all the dataset and got different types of accuracy and the accuracy was good. Models like CNN, SVM, Bengali-Net etc. were used.

Samrat Alam et al. [17] presented a topic named Bangla text categorization on Bangla news articles based on 12 categories. The set of data was extracted from different Bangla newspapers reports. Various models were used like CNN, LSTM, CNN+Bi-LSTM methods were applied on the dataset. Hybrid model gave the highest accuracy of 88.56% with ten categories and 84.93% on all categories that had been collected. The word embedding is used for feature extraction.

Amartya Roy et al. [18] preferred a topic named Bangla Text Classification with a new multiclass dataset. Using features class algorithms like NB, SVM, HMM, RF, BERT etc. There were 2 dataset and in the 1st dataset, SVM gave the highest accuracy of 92.61% and in the 2nd dataset XLM-RoBERTa-Large gave 95.13% accuracy. The second dataset had a total 12,695 documents and the second dataset had 1756 documents.

Kamrus Salehin et al. [19] came up with a topic of text classification techniques specially for Bangla news. Various types of ML models, also deep learning techniques were used. They are LSTM, SVM, XGBoost, MLP, LR, MNB, RF and among these algorithms LSTM gave 87%. Total samples were 75,951 and it was divided into 12 classes. The data was collected from news websites and blogging platforms.

Md. Mahbubur Rahman et al. [20] recommended Bengali documents classification using character level deep learning. CNN & LSTM were applied on the dataset. LSTM gave 95.42% in Prothom Alo dataset. Dataset were collected online via newspaper, ai tools etc. Almost 2 lakh data were used in this topic. In the research hybrid models might give more stable results.

Saifur Rahman et al. [21] presented Bangla documents classification using Deep recurrent neural networks with Bi-LSTM. Also, TF-IDF, Word2Vec, SVM, KNN, MNB were used on the Dataset. Total number of words in the documents were 16,889,327. Dataset was collected from news articles. RNN model obtained 98.33% accuracy.

## III. Methodology

### A. Workflow:

The translation is from Sylheti Bangla to Modern Bangla language using Natural Language Processing. Fig. 1. shows the total work flow. After the data collection, necessary preprocessing is used for cleaning the dataset. After preprocessing Translation is done, some Natural Language Processing models were applied and trained the best fit model with the dataset.

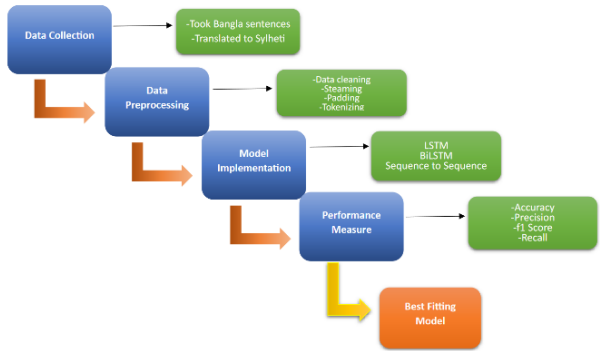

**Fig. 1.** Diagram of the workflow

### B. Dataset:
#### 1. Data Collection:

The data is collected from the citizens of Sylhet region. Dataset consists of a total 1200 Modern Bangla sentences that are translated to Sylheti Bangla language. Sylheti locally spoken Bangla is different from the actual Bangla. So, the dataset has two attributes Modern Bangla which is the actual Bangla language and Sylheti Bangla that are spoken in the Sylhet region of Bangladesh. Table I. shows few samples of the dataset.

**TABLE I.** SAMPLE DATASET

| Modern Bangla | Translated Sylheti Bangla |
|---|---|
| আমি তোমাকে শুনতে পাচ্ছি না | আমি তোমারে ছুনিয়ার না |
| রাগ করো না | রাগ কইর না |
| তার হাসি ভাল ছিল | হের আসি সুন্দর আছিল |
| আমি পড়াশোনা করতে যাচ্ছি | আমি পড়াত যাইরাম |
| বনের জন্তুদের মধ্যে হাতি সবচেয়ে বড় | বনের জন্তুর মাঝে হাতি হইল হক্কলের থাকি বড় |

#### 2. Data Preprocessing:

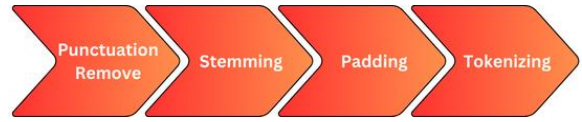

**Fig. 2.** Diagram of order of the data processing steps

It has been made sure that the collected data or the sentences are not consisting of any Non character letter, non-Bengali character or any kind of duplicate values. 80% data used for training the model and the rest of the 20% used in testing the trained system. Fig. 2. Shows the data preprocessing steps.

#### 2.1. Punctuation Remove:

At first, all punctuations are removed for betterment in training the models. This preprocessing step helped to optimize the models even more to understand the train data. Some Bangla punctuations are, ',' '।' '?' '!'

#### 2.2. Stemming:

This is an important preprocessing step. Stemming changes a word to its founding or root form. It reduces the dimension of a word. Bangla is a language whose resources are very low. As a result, there aren't enough good stemmers.

#### 2.3. Padding:

The padding maintains a sequence so that all sentences have the same length. This is an important preprocessing step. Padding is usually done by adding tokens.

#### 2.4. Tokenization:

The preprocessing step is the most important part of the preprocessing steps. It converts raw text data into a style that is easy to analyze for the models. Table II. shows the example of tokenization.

**TABLE II.** TOKENIZING EXAMPLE

|  | Before Tokenizing | After Tokenizing |
|---|---|---|
| Actual Bangla | দয়া করে সব কথা খুলে বলুন | 'দয়া', 'খরইন', 'তাইলে', 'খতা', 'খুলিয়া', 'খইন' |
| Sylheti | দয়া কইরা সব খতা খুলিয়া খও আমারে | 'দয়া', 'করুন', 'কথা', 'খুলে', 'বলুন' |

After all the preprocessing steps the dataset is ready for training models. It required vectorization for making the data understandable for the machine. After all the preprocessing methods finally the data ready for training and split in two portions.

**C. Model Description**

The main goal of this research is to translate from Local Sylheti Bangla to Modern Bangla which is known by all. For the translation, different deep learning models like LSTM, Bi-LSTM, Seq2Seq models. These models are very popular in NLP or Natural language processing research.

**LSTM (Long Short-Term memory)**

The model is known as recurrent neural network architecture. The model is drafted to detect historical information of time series data and analyze sequential data such as speech, text etc. The LSTM model is mainly used in AI, deep learning and is very popular in NLP research. Also, it's used to design vanishing gradient problems. This problem is generally found in RNN. The vocabulary size is set to 10000 and maximum length to 50. Each layer will accept about 50-dimensional input. Also, there are some layers known as embedding 11 and embedding 12. The embedding dimension is 64. The layers convert input tokens into dense vectors and their size is 64. LSTM takes input form embedding 12. The layer is connected with a hidden state and a final cell. The outer space layer of our model is 128 which is also known as LSTM units. There is only 1 dense layer and the output shape is (None,50,10000) and contains 2,767,632 parameters. Each size of a vector element is 10,000. 50 epochs were used with 32 batch sizes.

$$i_t = \sigma(W_i[h_{t-1}, x_t] + b_i) \quad (2)$$

$$\tilde{C}_t = tanh(W_{\tilde{C}}[h_{t-1}, x_t] + b_{\tilde{C}}) \quad (3)$$

$$f_t = \sigma(W_f[h_{t-1}, x_t] + b_f) \quad (4)$$

$$o_t = \sigma(W_o[h_{t-1}, x_t] + b_o) \quad (5)$$

$$h_t = o_t * tanh(C_t) \quad (6)$$

Fig. 3. there are input gates, output gates, forget gate and call state update. The forget gate is defined with a sigmoid function and tells what information will be shown from the previous state. The input gate is also defined with a sigmoid function but it indicates new information stored. And the output gates decide what part will be shown as output.

**Bi-LSTM (Bidirectional LSTM)**

Bi-LSTM is an extended version of the LSTM neural network model. It processes input orders in both backward direction and forward direction. It's also a recurrent neural network model. It's a sequence processing model for both directions that consists of two LSTMs, one takes the input in forward direction and one takes input for backward direction. The model can both store past and future context for every step. This process makes the model more effective. Bi LSTM understands entity recognition, sentiment analysis etc. In the model summary it could be found or detected some interesting things. There are several layers in the model architecture. In the embedding layer, there are a total 317,440 parameters. The input is converted into integer indices into dense vectors and its size is 256 in the model for each word embedding. Output shape shows that this model can handle up to 15 words. Then the bidirectional layer contains 3,149.824 parameters. Input sequence of the sentences can be processed into both forward and backward. For each word the output shape represents the word sequence into a vector size of 1024 for both past and future predictions. The attention layer contains 1,039 parameters. The input sequence calculates weight when predicting something. Output layer doesn't change in this process. There are a total of 524,800 parameters. The input data uses linear transformation and makes it into a 512 lower dimensional space. No parameters were used in the dropout layer. This layer applies certain techniques to prevent overfitting. The final layer or the dense layer contains 636,120 parameters and it generates the model output. The model used in this work gives a vector size of 1240. Also, there is a custom attention layer added. The layer is used to get a good optimal result and it helps the model to focus on the necessary or relevant parts to get a good and optimal result. 50 epochs were used without early stopping with a batch size of 64. Total params are 14,683,595 and optimized params are 9,789,064. Trainable params are 4,894,531.

$$\vec{h}_t = LSTM_{forward}(x_t, \vec{h}_{t-1}) \quad (7)$$

$$\vec{h}_t = LSTM_{backward}(x_t, \vec{h}_{t-1}) \quad (8)$$

Fig. 4. shows the input, forward layer, backward layer, outputs and connecting arrows. The input passes a sequence in the architecture. Forward layers and backward layers contain LSTM cells. For forward the sequence is in forward direction and for backward the sequence is in backward direction. The output is represented with a sigmoid function in yellow color. The output is basically the combination of forward and backward LSTM.

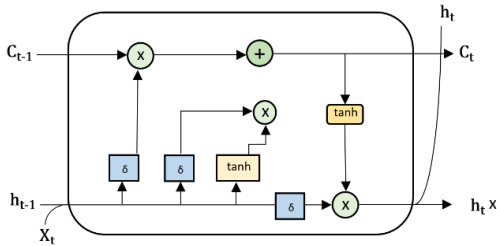

**Fig. 3.** LSTM model architecture contains interacting layers

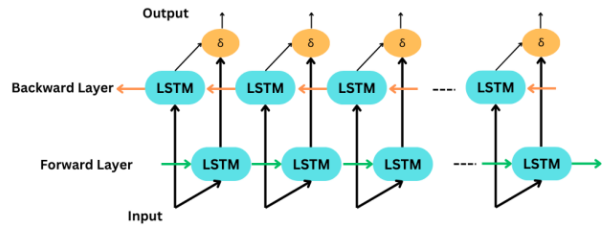

**Fig. 4.** Bi-LSTM model architecture

**Seq2Seq (Sequence to Sequence)**

The Sequence to Sequence model is a machine learning model and uses transformation to change the sequence of a process. The model contains both encoder and decoder. The encoder normalizes the input into a fixed size vector. The decoder gives an output sequence based on the input vector. This model is usually used in NLP tasks. Language translation, conversational models, Text summarization, time series forecasting. There are several layers in this model too. First of all, input layers, two input layers but number of params are 0. But different output shapes of two different layers inside the input layer. The input layers show the shapes of input data. The first input layer expects sequence length of 12 and accordingly the second input layer expects sequence length of 11. Next the embedding layer, in fact there are two embedding layers. Embedding is connected with the input layer. The Number of params is 317,440. The embedding 1 is connected with input layer 1 with 226,560 parameters. Both of them got different output shapes. These layers convert mainly integer encoded into dense vectors and the fixed size is 256 for this research. The output shape represents the token as a 256-dimensional vector. And then there are two LSTM layers and both of them have different parameters and definitely the output shapes are different too. LSTM is connected to embedding and LSTM 1 is connected with 3 layers, the names of those layers are: embedding 1, LSTM 0,1 and LSTM 0,2. The first LSTM layer and the second layer processes the sequence of input into 12 and 11. The hidden state size for both is 256. Finally, the dense layer is connected with lstm_1 0,0 with 227,445 parameters. This applies linear transformation for the final output and the output shape is 885 vectors for each token in the given sentence. The max length is 12 and the latent dimension is 256, with 32 batch size and 50 epochs. Total number of params and trainable params are the same and their number are 2,054,029.

In Fig.5. there are input, encoder, state, decoder and output. Input sequence that is given into the encoder. The encoder summarizes the data into fixed size context vectors. This state vector contains the summarization and passes it to the decoder for the next step. It gives output in sequence to sequence and it produces output tokens. The decoder outputs the final result. Finally, the output just shows it.

models are built with Keras, tensor flow and python. A free cloud named Jupyter Notebook was used to execute the codes. To calculate the accuracy, f1 score, precision and recall this formula were applied:

*Accuracy:* The accuracy of a model refers to the proportion of predictions it makes calculated as the ratio of positives and true negatives, to all positive and negative observations.

$$Accuracy = \frac{TP+TN}{TP+TN+FP+FN} \quad (9)$$

*F1 Score:* A weighted average of precision and recall, where the weights are equal. It's a more comprehensive evaluation metric than accuracy because it maximizes two competing objectives simultaneously. F1-score is the harmonic mean of precision and recall score.

$$f1\ score = \frac{2*Precision*Recall}{Precision+Recall} \quad (10)$$

*Recall:* The proportion of true positive predictions among all the actual positive samples in the dataset. It's also known as sensitivity and focuses on the model's ability to identify all positive instances.

$$Recall = \frac{TP}{TP+FN} \quad (11)$$

*Precision:* The proportion of true positive predictions among all the positive predictions made by the model. It focuses on the accuracy of positive predictions.

$$Precision = \frac{TP}{TP+FP} \quad (12)$$

Some deep learning models like LSTM, Bi-LSTM, Seq2Seq models were applied in the research. Fig. 6. shows the accuracy of different models used in the research are 89.7%, 76.2% and 70.7% are showed in bar chart. Also, the model names were specified.

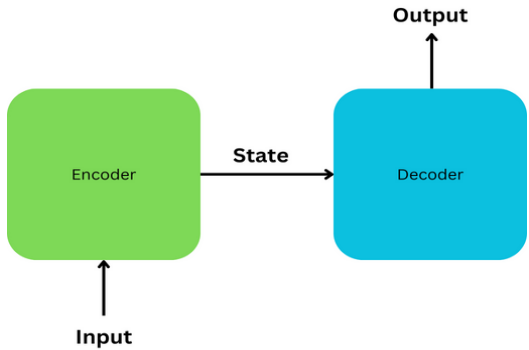

**Fig. 5.** Sequence to Sequence model architecture

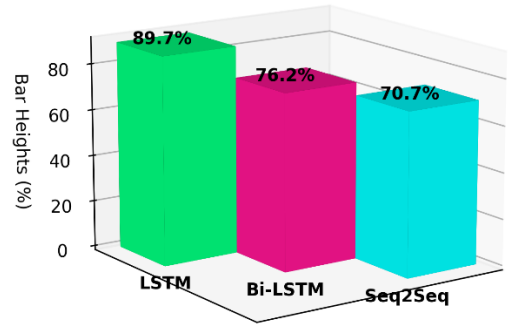

**Fig. 6.** Accuracy comparison bar chart of 3 models used

## IV. Result Analysis & Discussion

The research was conducted using 1200 sentences and contains two columns named 'Bangla' and 'Sylheti'. The

**TABLE III:** CLASSIFICATION TABLE

| Models | Accuracy | Precision | Recall | F1-Score |
|---|---|---|---|---|
| LSTM | 0.8973 | 0.8745 | 0.8973 | 0.8838 |
| Bi-LSTM | 0.7625 | 0.7203 | 0.7625 | 0.7392 |
| Seq2Seq | 0.7069 | 0.69 | 0.7069 | 0.6915 |

Table III. shows there is the classification report of the three models. The research could be more optimal if the dataset was larger than our raw dataset. Because there was a higher chance that our models might give greater accuracy if the dataset was larger. No model can predict 100% correctly. There are some barriers for each model. Some models can predict quite well but some models can't predict well. Because the environment or dataset isn't suitable for the model, that's why some models can't perform up to our expectations.

Fig. 7. Fig. 8. & Fig. 9. Shows the train and validation accuracy plots are shown for 3 different models. For LSTM, the train and validation accuracy difference are very low and approximately after 40 epoch train accuracy got higher than the validation accuracy. For Bi-LSTM, up-to 30 epochs the train and validation accuracy graduation went up but after 40 epochs, the training accuracy went up at a good rate. For sequence to sequence, the train accuracy went up higher but validation accuracy went up and the growth is slower.

Fig. 10. Fig. 11. & Fig. 12. shows the training and validation loss of the applied models. The training and validation loss are almost the same. The loss for LSTM for train and validation is almost same but the loss of train is little-bit less than the validation loss. For Bi-LSTM, the train and validation lost was almost same up-to 13 epochs, after that the train loss went down but the validation loss went up. The sequence to sequence training loss is going down after around 18 epochs but the validation loss is almost the same till the end.

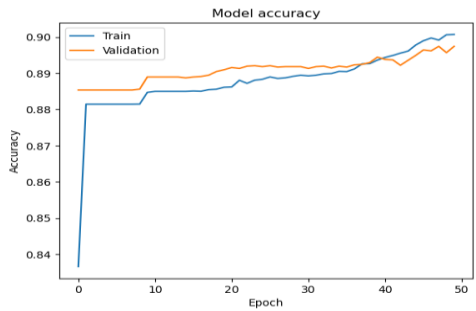

**Fig. 7.** Train and Validation Accuracy Plot for LSTM

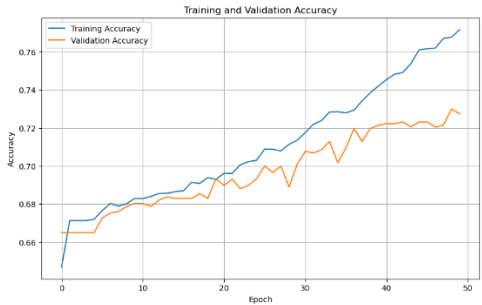

**Fig. 8.** Train and Validation Accuracy Plot for Bi-LSTM

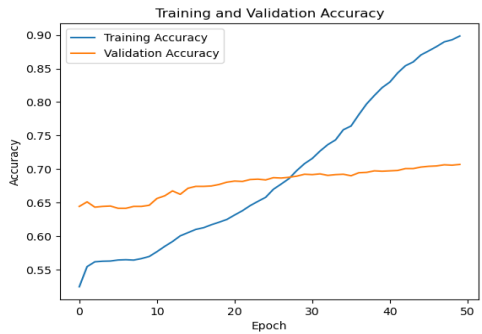

**Fig. 9.** Train and Validation Accuracy Plot for Sequence to sequence

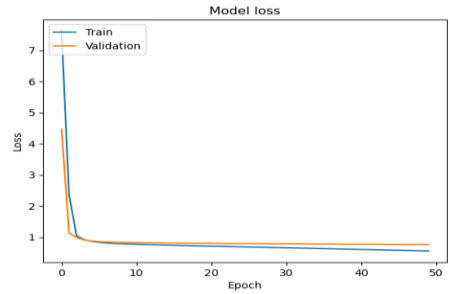

**Fig. 10.** Train and Validation Loss Plots for LSTM

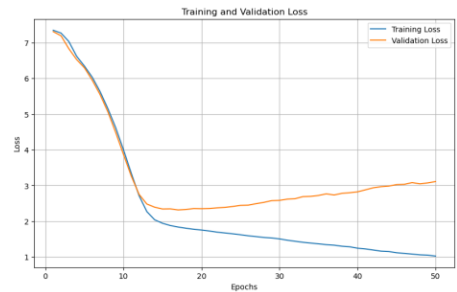

**Fig. 11.** Train and Validation Loss Plots for Bi-LSTM

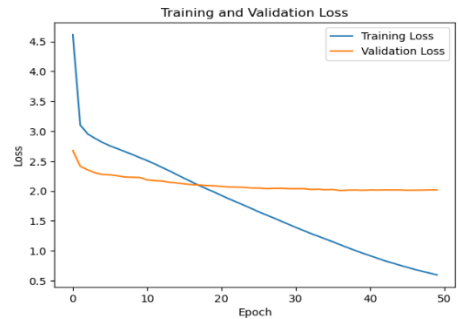

**Fig. 12.** Train & Validation Loss Plots for sequence to sequence

## V. Conclusion & Future Work

The conclusion is that, even though Bangla being the national language of Bangladesh, people from different areas speak different words and sentences and these are quite difficult for a person to understand a different division's language specially Sylheti Bangla sentences. This research offered a pipe line that translates Sylheti sentences to pure Bengali sentences. In this work 1200 sentences are taken and translated to Sylheti. After preprocessing the model is trained using this dataset. The models that are selected, are popular among Bangla NLP researchers. A lot of research were done using the models, LSTM, Bi-LSTM and Sequence to Sequence, these three models were trained in the research. As a result, these models were used in this research. LSTM performed better on this dataset and scored 89.3% accuracy. This work has a better future scheme for Bengali NLP researchers. There is a future plan on this paper to broaden the scope and include other regions' languages too and build an application that will provide the service. Also in addition, the total data used in this work is low, it's a reason that the objective is to work with this project with a big size of dataset. In future, if Hybrid models were applied here the results or the accuracy could have been different or might achieve higher accuracy than this. Also, there are not good stemmers for Bangla, in future good and optimized stemmers are made then this type of research might get very popular among the researchers.

## Reference

[1] Siddique, S., Ahmed, T., Talukder, M.R.A. and Uddin, M.M., 2021. English to bangla machine translation using recurrent neural network. *arXiv preprint arXiv:2106.07225*.

[2] Winata, G.I., Aji, A.F., Cahyawijaya, S., Mahendra, R., Koto, F., Romadhony, A., Kurniawan, K., Moeljadi, D., Prasojo, R.E., Fung, P. and Baldwin, T., 2022. NusaX: Multilingual parallel sentiment dataset for 10 Indonesian local languages. *arXiv preprint arXiv:2205.15960*.

[3] Roy, A., Dhar, A.C., Akhand, M.A.H. and Kamal, M.A.S., 2021. Bangla-english neural machine translation with bidirectional long short-term memory and back translation. *Int. J. Comput. Vis. Signal Process*, 11(1), pp.25-31.

[4] Sarkar, Andkon, Aishwarja Paul Sourav, and Rezvi Ahmed. "Sentiment analysis in Bengali Text using NLP." PhD diss., Brac University, 2023.

[5] Ahmed, Md Tofael, et al. "Natural language processing and machine learning based cyberbullying detection for Bangla and Romanized Bangla texts." TELKOMNIKA (Telecommunication Computing Electronics and Control) 20.1 (2021): 89-97.

[6] Wadud, M.A.H., Mridha, M.F. and Rahman, M.M., 2022. Word embedding methods for word representation in deep learning for natural language processing. *Iraqi Journal of Science*, pp.1349-1361.

[7] Hossain, M.Z., Rahman, M.A., Islam, M.S. and Kar, S., 2020. Banfakenews: A dataset for detecting fake news in bangla. *arXiv preprint arXiv:2004.08789*.

[8] Fouad, K.M., Sabbeh, S.F. and Medhat, W., 2022. Arabic Fake News Detection Using Deep Learning. *Computers, Materials & Continua*, 71(2).

[9] Bhowmik, N.R., Arifuzzaman, M. and Mondal, M.R.H., 2022. Sentiment analysis on Bangla text using extended lexicon dictionary and deep learning algorithms. *Array*, 13, p.100123.

[10] Das, R.K., Islam, M. and Khushbu, S.A., 2023. BTSD: A curated transformation of sentence dataset for text classification in Bangla language. *Data in Brief*, 50, p.109445.

[11] Bhowmik, N.R., Arifuzzaman, M., Mondal, M.R.H. and Islam, M.S., 2021. Bangla text sentiment analysis using supervised machine learning with extended lexicon dictionary. *Natural Language Processing Research*, 1(3-4), pp.34-45.

[12] Sayeed, A., Shin, J., Hasan, M.A.M., Srizon, A.Y. and Hasan, M.M., 2021. Bengalinet: A low-cost novel convolutional neural network for bengali handwritten characters recognition. *Applied Sciences*, 11(15), p.6845.

[13] Roy, A., Sarkar, K. and Mandal, C.K., 2023. Bengali Text Classification: A New multi-class Dataset and Performance Evaluation of Machine Learning and Deep Learning Models.

[14] Rahman, S. and Chakraborty, P., 2021, May. Bangla document classification using deep recurrent neural network with BiLSTM. In *Proceedings of International Conference on Machine Intelligence and Data Science Applications: MIDAS 2020* (pp. 507-519). Singapore: Springer Singapore.

[15] Hussain, M.G., Hasan, M.R., Rahman, M., Protim, J. and Al Hasan, S., 2020, August. Detection of bangla fake news using mnb and svm classifier. In 2020 International Conference on Computing, Electronics & Communications Engineering (iCCECE) (pp. 81-85). IEEE.

[16] Mugdha, S.B.S., Ferdous, S.M. and Fahmin, A., 2020, December. Evaluating machine learning algorithms for bengali fake news detection. In 2020 23rd International Conference on Computer and Information Technology (ICCIT) (pp. 1-6). IEEE.

[17] Wijonarko, P. and Zahra, A., 2022. Spoken language identification on 4 Indonesian local languages using deep learning. *Bulletin of Electrical Engineering and Informatics*, 11(6), pp.3288-3293.

[18] Sen, O. and Roy, P., 2021, September. A convolutional neural network based approach to recognize Bangla spoken digits from speech signal. In *2021 International Conference on Electronics, Communications and Information Technology (ICECIT)* (pp. 1-4). IEEE.

[19] Hasan, M., Puja, S.P., Bijoy, M.H.I., Sattar, A. and Rahman, M.M., 2022, October. Multiple Bangla Sentence Classification using Machine Learning and Deep Learning Algorithms. In *2022 13th International Conference on Computing Communication and Networking Technologies (ICCCNT)* (pp. 1-6). IEEE.

[20] Alam, S., Haque, M.A.U. and Rahman, A., 2022, February. Bengali text categorization based on deep hybrid CNN–LSTM network with word embedding. In *2022 International Conference on Innovations in Science, Engineering and Technology (ICISET)* (pp. 577-582). IEEE.

[21] Salehin, K., Alam, M.K., Nabi, M.A., Ahmed, F. and Ashraf, F.B., 2021, December. A comparative study of different text classification approaches for bangla news classification. In *2021 24th International Conference on Computer and Information Technology (ICCIT)* (pp. 1-6). IEEE.